\documentclass{svjour3} 
\smartqed 
\usepackage{graphicx}
 \usepackage{lineno,hyperref}
\usepackage{textcomp}
\usepackage{graphicx}
\usepackage{caption}
\usepackage{url}
\usepackage[usenames, dvipsnames]{color}
\usepackage{float}
\begin{document}

\title{New Graph-based Features For Shape Recognition}


\author{Narges Mirehi \and
Maryam Tahmasbi \and Alireza Tavakoli Targhi 
}
\institute{ Maryam Tahmasbi \at
Department of computer science, Shahid Beheshti University, G.C., Tehran, Iran.\\
Tel.: +98 21 2990 3004\\
Fax: +98 21 2243 1653\\
\email{m\_tahmasbi@sbu.ac.ir} \\
\and
Narges Mirehi
\at
Department of computer science, Shahid Beheshti University, G.C., Tehran, Iran.\\
\email{n\_mirehi@sbu.ac.ir}\\
\and \\
Alireza Tavakoli Targhi
\at
Department of computer science, Shahid Beheshti University, G.C., Tehran, Iran.\\
\email{a\_tavakoli@sbu.ac.ir}
\and \\
}


\maketitle

\begin{abstract}

Shape recognition is a main challenging problem in computer vision.
Different approaches and tools are used to solve this problem.
Most existing approaches to object recognition are based on pixels. Pixel-based methods are dependent on the geometry and nature of the pixels, so the destruction of pixels reduces their performance.
In this paper, we study the ability of graphs as shape recognition. We construct a graph that captures the topological and geometrical properties of the object. Then, using the coordinate and relation of its vertices, we extract features that are robust to noise, rotation, scale variation, and articulation. To evaluate our method, we provide different comparisons with state-of-the-art results on various known benchmarks, including Kimia\textquotesingle s, Tari56, Tetrapod and Articulated dataset. We provide the analysis of our method against different variations. The results confirm our performance, especially against noise.


\end{abstract}

\begin{acknowledgements}
 We would like to to thank the helpful comments and
suggestions from Dr Sven Dickinson.
\end{acknowledgements}

\bibliographystyle{plain} 


\section{Introduction}
Shape is a significant concept in image understanding. It includes considerable meaningful properties of an object
and provides stability to different object deformations (such as articulation, occlusion, and noise) and transformations ( such as rotation, translation, scale and etc.).
Therefore the development of object descriptors which include shape information is an essential problem. For this purpose, different approaches are presented such as skeletal, geometrical and graph-based methods \cite{Belongie2007,Sebastian2004,Ling,Wang2012,Yang2016,YangInvariant2016,Jia2016}.
The skeleton is a connected set of medial lines inside the shape along the limbs that captures the shape of the boundaries.
Some proposed methods utilize the skeleton for object recognition \cite{Macrini2011,Siddiqi1999,Yang2016}. Skeleton-based methods are sensitive to noise on the boundary. In fact, little variation or noise in the object can cause considerable deformations and redundant branches in the skeleton and its topology might be disturbed.
Siddiqi et al. presented shock graph which represents skeleton information of an object in a directed acyclic graph \cite{Siddiqi1999}. It encodes the local variation of the radius function along the medial axis in the nodes of the graph.
Macrini et al. introduced bone graph that improved stability over shock graph. Its nodes represent the non-ligature segments of the medial axis and its edges show the ligature segments that capture the relational information between medial parts \cite{Macrini2011}.

 Yang et al. suggested a hierarchical skeleton-based algorithm to increase the stability of skeleton pruning which captures different levels of skeletons \cite{Yang2016}.

 Some interesting geometric and graph-based approaches in shape recognition include shape context ~\cite{Belongie2007}, inner-distance \cite{Ling}, graph edit distance \cite{Gao2010}, triangle area representation~\cite{Alajlan2007}, height function~\cite{Wang2012}, invariant multi-scale \cite{YangInvariant2016} and hierarchical characteristic number contexts (HCNC) \cite{Jia2016}.\\
The major disadvantage of geometric features is sensitivity to articulation and some deformations. Belongie defined a shape context (SC) descriptor which for every contour point, stores the distribution of the rest of the points with respect to it. The shape context of every point is a histogram of relative coordinates of remaining points in log-polar space, which leads to more sensitivity of the descriptor to nearby sample points than to farther points \cite{Belongie2007}.
Alajlan proposed triangle area representation (TAR) to measure the convexity and concavity of boundary points.
The triangles are formed by boundary points in different sale levels and their area of a boundary point is signed by positive, negative or zero in convex, concave or straight line position respectively \cite{Alajlan2007}.
Gao et al. suggested graph edit distance (GED) which measures the similarity between pairwise graphs in inexact graph matching. GED is defined as the cost of the least sequence of edit operations needed to transform a graph into another \cite{Gao2010}.
\\
Payet and Todorovic introduced a hierarchical graph to describe an object. The nodes of this graph include the geometric properties of corresponding parts, and the edges store the strength of neighbor and interactions between the parts. Their matching algorithm finds a subgraph isomorphism of the minimum cost \cite{Payet2009}.

Ling and Jacobs introduced the inner-distance shape context (IDSC) approach \cite{Ling}. To compute the inner-distance of a shape, a graph is constructed; its vertices are the contour points and the edges are drawn between points that their connecting segment lies inside the shape.
The length of the edges equals the Euclidean distance between nodes, so does the distance between different contour points and the shortest path between them in this graph.\\
Another geometric method was proposed based on height functions \cite{Wang2012}.
The method considers a fixed number of points on the boundary of an object and for each point measures the height of other points from its tangent line. The shape is described using smoothing the height functions. Nevertheless, height functions can be sensitive to articulation and the feature vector for each contour point is long.\\
Yang et al. proposed an invariant multi-scale method to capture local and global data of the object \cite{YangInvariant2016}. This method considers circles with different radii on each boundary point and measures features such as area, arc length and central distance of boundary points inside these circles. The similarity between objects is calculated via dynamic programming algorithm.\\
Jia et al. introduced Hierarchical projective invariant contexts (HCNC) method by computing characteristic number values on a
series of 5 sample points on the object contour \cite{Jia2016}. HCNC is based on local features and is sensitive to noise.

Different graph-based methods were presented in recent decade \cite{Gao2010,Payet2009}. Most of these methods store intended features in nodes and edges of a graph and use this graph for efficient object matching. In fact, the role of graph in matching is more important than that of representation \cite{Gao2010}.

 All mentioned approaches have some limitations and suffer sensitivity to noise, articulation and some deformations. 
 In this paper, we study object recognition from graph theory viewpoints. Graphs are robust with respect to rotation, articulation, and noise, so they can be effective tools to capture the image properties. %
  These graphs have limited number of vertices and make the size of the problem fixed in different scales. So, it can be used as powerful tools in shape recognition.
We use Growing Neural Gas (GNG) algorithm \cite{Fritzke1995} to construct the graph. This algorithm constructs a GNG graph model of input data incrementally.

Two principal properties of this graph are low dimensionality and topological preservation, i.e. the number of vertices of the graph does not depend on image scale and it is not sensitive to articulation and noise on the boundary. In this graph, every vertex has a coordinate, so we can use geometric properties of the graph as well.
 Also, in other research, GNG graph is applied to hand gesture recognition  \cite{mirehi2019hand}.
 We use both topological and geometrical properties of this graph to extract meaningful features from the image. Both theoretic discussions and experimental results show that our method is invariant to articulation, noise, occlusion, rotation, and scale. To evaluate our method, we compare our results on various challenging benchmarks that include different variations.

The rest of this paper is organized as follow: we introduce our method, construct the GNG graph and extract its outer boundary in section \ref{sec:methodology}, then we define the features in section \ref{sec:feature extraction} and the matching algorithm in section \ref{sec:Similarity measureing}, and in section \ref{sec:Comparison}, we compare the results and evaluate our approach. Finally, we present the conclusion and open problems in section \ref{sec:conclusion}.

\section{Methodology}\label{sec:methodology}
In this section, we introduce our proposed method. We use a graph to approximate the object. The vertices are scattered almost uniformly inside the object. The main steps of our proposed method are:
\begin{enumerate}
\item Constructing a graph that models the object useing Growing Neural Gas (GNG) algorithm \cite{Fritzke1995}. The constructed graph is called \textit{GNG graph}.
\item Extracting the outer boundary of the GNG graph, using computational
geometry approaches and extracting geometrical and topological features from this graph.
\item Measuring the similarity between objects using dynamic programming algorithm.
\end{enumerate}
\vspace{-0.5cm}
\subsection{Why GNG graph}
Our method is based on a graph with vertices uniformly placed inside the image. This graph must have the following properties:
 \begin{itemize}
\item The vertices are placed almost uniformly inside the object and the edges have almost equal lengths.
\item The number of vertices is fixed and does not depend on the scale of the object.
\item The graph is robust with respect to noises. It must ignore the holes and cracks inside and the noise on the boundary of the object.
\end{itemize}

Different approaches can be chosen to construct this graph. We choose GNG algorithm because it well satisfies the mentioned properties, the running time is satisfiable and can be extended to 3d object representation and object tracking \cite{Holdstein2008,Escolano2016,Sun2017}. Also, since the structure is not fixed but adaptable, GNG can learn new evolving patterns in an online learning process and is able to adapt to dynamic changing operating conditions \cite{Fink2015}.

 \subsection{Constructing the GNG graph} \label{sec:GNG}
Growing neural gas algorithm (GNG) is an unsupervised incremental algorithm which learns the topology of the input data utilizing competitive Hebbian learning \cite{Fritzke1995}. It fills the area of an object by vertices almost uniformly distributed inside it and describes the distribution of input data using less space. The output of the algorithm is a graph preserving the topological structure of input data.
The GNG algorithm is presented in the following \cite{Fritzke1995}:\\
\\
\textbf{GNG algorithm}
\begin{enumerate}
\item
Start the GNG graph with two neurons $a$ and $b$ in random positions $w_a$ and $w_b$.
\item
Generate a new input $x$ of input space.
\item
Find the nearest node $s_1$ and the second nearest node $s_2$ to the input vector $x$.
\item
Increase the age of all adjacent edges of $s_1$. (every edge includes a parameter of age with the initial value of zero).
\item
Increase the error variable of $s_1$ by the square of the Euclidean distance between it and the input vector $x$ (Every node $i$ has an error variable $e_i$ which is initialized as zero).
$$\Delta error_{s_1}={\parallel w_{s_1 }-x \parallel }^2$$
\item
Move the location of $s_1$ and its neighbors (n) with a multiple of $\epsilon_b$ and $\epsilon_n$ toward input vector, respectively.
$$\Delta w_{s_1 }=\varepsilon_b (x-w_{s_1 })$$
$$\Delta w_n=\varepsilon_n (x-w_{n })$$
\item
If $s_1$ and $s_2$ are connected by an edge, set the age of this edge to
zero or else add an edge with age zero between $s_1$ and $s_2$.
\item
Remove all edges with age larger than $\alpha_{max}$.
\item
When the comparison is finished, insert new node $r$ between a node $q$ with maximum error variable and its neighbor $f$ which has the largest value of error variable.
$$ w_r=(w_q+w_f)/2$$
\begin{itemize}
\item
Insert edges connecting $r$ with $q$ and $ f$, and remove the original edge between $q$ and $f$.
\item
Decrease the error variables of $q$ and $f$ by multiplying with a constant $\alpha$.
\end{itemize}
\item
Decrease all error variables by multiplying them by a constant $\sigma$.
\item
If the stopping criterion (e.g, the number of inputs is a multiple of a parameter $\lambda$) is
not yet fulfilled, go to Step 2.\\
\end{enumerate}
\begin{center}
\begin{figure*}[htb]
\vspace*{2cm}
\centering
\includegraphics[scale=.3]{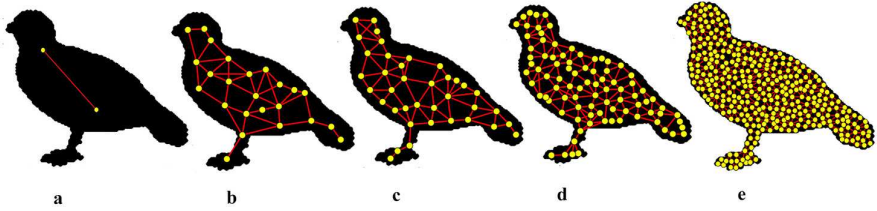}
\caption{\label{fig:gng}The different steps of GNG algorithm on a binary image of Kimia \textquotesingle s 216.}

\end{figure*}
\end{center}
 Figure \ref{fig:gng} shows a sample image and its GNG graph. To construct the GNG graph, the parameters must be chosen in a way that a more accurate graph results with less computation time. We tested different parameters and let $N=350 $, $\lambda=50$, $\alpha_{max}=50$, $\alpha =0.5$, $\sigma =0.995$, $\epsilon_b = 0.05 $, $\epsilon_n = 0.005 $, where $N$ is the number of neurons. We exprimented GNG graph with various number of neurons from ${150,200, 250,300,350}$ and observed that 350 neurons are sufficient. After applying the GNG algorithm, some corrections might still be needed.
\begin{itemize}
\item Some edges in the GNG graph are redundant, it means that they go
through the background. We find and remove these edges from the
graph. In order to do so, we find the middle point of each edge and
consider its nine neighbors, if most of them belong to the
background, we remove the edge.
\item We suppose that the GNG graph is connected. If not, we choose
the connected component with the largest number of nodes.
Disconnectivity happens in the presence of noise, or multiple objects are formed in the image. We suppose that the image contains only one object.
\end{itemize}

 \subsection{Extracting the outer boundary of the GNG graph}\label{sec:outer-boundary}
In this phase, we extract the outer boundary of the GNG graph. If the graph is 2-connected, then the outer boundary is a cycle, otherwise, it is a closed walk. So, we can store its vertices in a cyclic array, $C$, in clockwise order of appearance. The idea is similar to the idea of a convex hull algorithm \cite{DeBerg2008}.\\

\begin{figure*}[!htb]
\includegraphics[scale=.45]{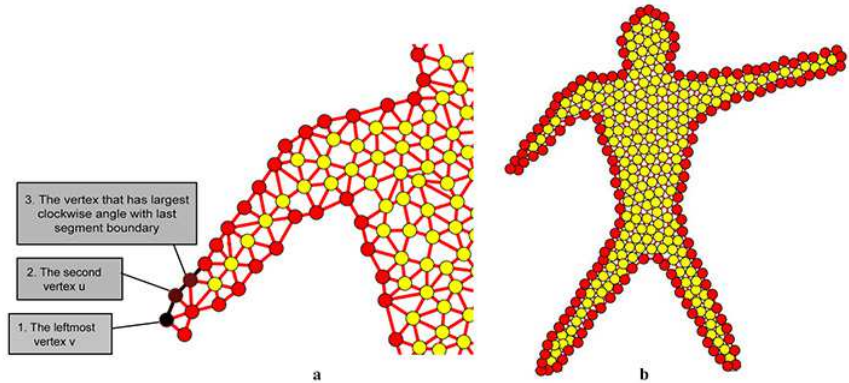}
\centering
\caption{\label{fig:boundary} a) The outer boundary extraction b) red vertices show the outer boundary of the graph.}
\end{figure*}

 \textbf{Outer boundary extraction}
\begin{enumerate}
\item
Find the leftmost vertex $v$ and its neighbor $u$ with the smallest clockwise angle with the vertical half-line crossing $v$.
\item
Insert $v$ and $u$ in $C$.
\item
Consider the two last vertices $i$ and $i-1$ in $C$ and for all vertices $j$, adjacent to $i$, compute the size of the clockwise angle at $i$ between the edges $i,i-1$ and $i,j$.
\item
Select a vertex with minimum angle as the next vertex on the boundary and insert it in $C$.
\item
If the last two vertices in $C$ are equal to initial vertices ($u$ and $v$), exit otherwise go to step 3. (This helps passing the cut vertex, if exists.)
\end{enumerate}
Figure \ref{fig:boundary} shows an example of a GNG graph and its outer boundary.

\section{Feature extraction} \label{sec:feature extraction}
Although the vertices on the outer boundary of GNG graph capture the special information of the object, the relation between boundary and internal vertices can provide new topological features that lead to better recognition. We introduce features for each vertex on the outer boundary, then compute and combine these features to describe global and local properties of the shape.

 The features are: \textit{perimeter $(P)$}, \textit{ boundary-in-disk $(B)$}, \textit{convex hull-area $(CH)$} and \textit{distance-to-center $(C)$} for outer boundary vertices of GNG.

 Given an object, let $U=\lbrace u_1,u_2,...,u_n \rbrace$ define the sequence of outer boundary vertices in clockwise order.
A feature vector $F_i=(P_i, B_i, CH_i, C_i)$ is computed for every outer boundary vertex $u_i$, $i \in \lbrace 1,...,n \rbrace $. $F_i$ consists of four features $P_i$, $B_i$, $CH_i$ and $C_i$ with lengths $\vert P_i \vert =m_1, \vert B_i\vert =m_2, \vert CH_i \vert =m_3, \vert C_i\vert =m_4 $ and $\sum_{j=1}^{4}m_j=m$. So $F_i$ can be considered as a $m$ dimensional vector.
We compute $F_i$ for every outer boundary vertex $u_i$, $i \in \lbrace 1,...,n \rbrace $ and define $F=F(U)=(F_1,F_2,F_3,...,F_n)$ which is a $m\times n$ matrix.\\
Let $G$ be the GNG graph of an object and $V(G)$ be the set of vertices in $G$. For every two vertices $v,u\in V(G)$, $d(u,v)$ shows their distance in $G$.

For every outer boundary vertex $u_i$ and for every integer $j$, let $D_j(u_i)=\{v\in V(G):d(v,u_i)\leq j \}$ be the discrete disk of radius $j$ around $u_i$ and $\bar{D}_j(u_i)=\{v\in V(G):d(v,u_i)=j \}$ be the boundary of the discrete disk. Figure \ref{fig:scale_gng} shows the discrete discs around a vertex in a sample image.
 We define four features $P$, $ B$, $CH$ and $C$ for a given object $A$ with outer boundary vertices $U=\lbrace u_1,u_2,...,u_n \rbrace$ as follows:
 \begin{itemize}

 \begin{figure}[!htb]
\includegraphics[scale=.4]{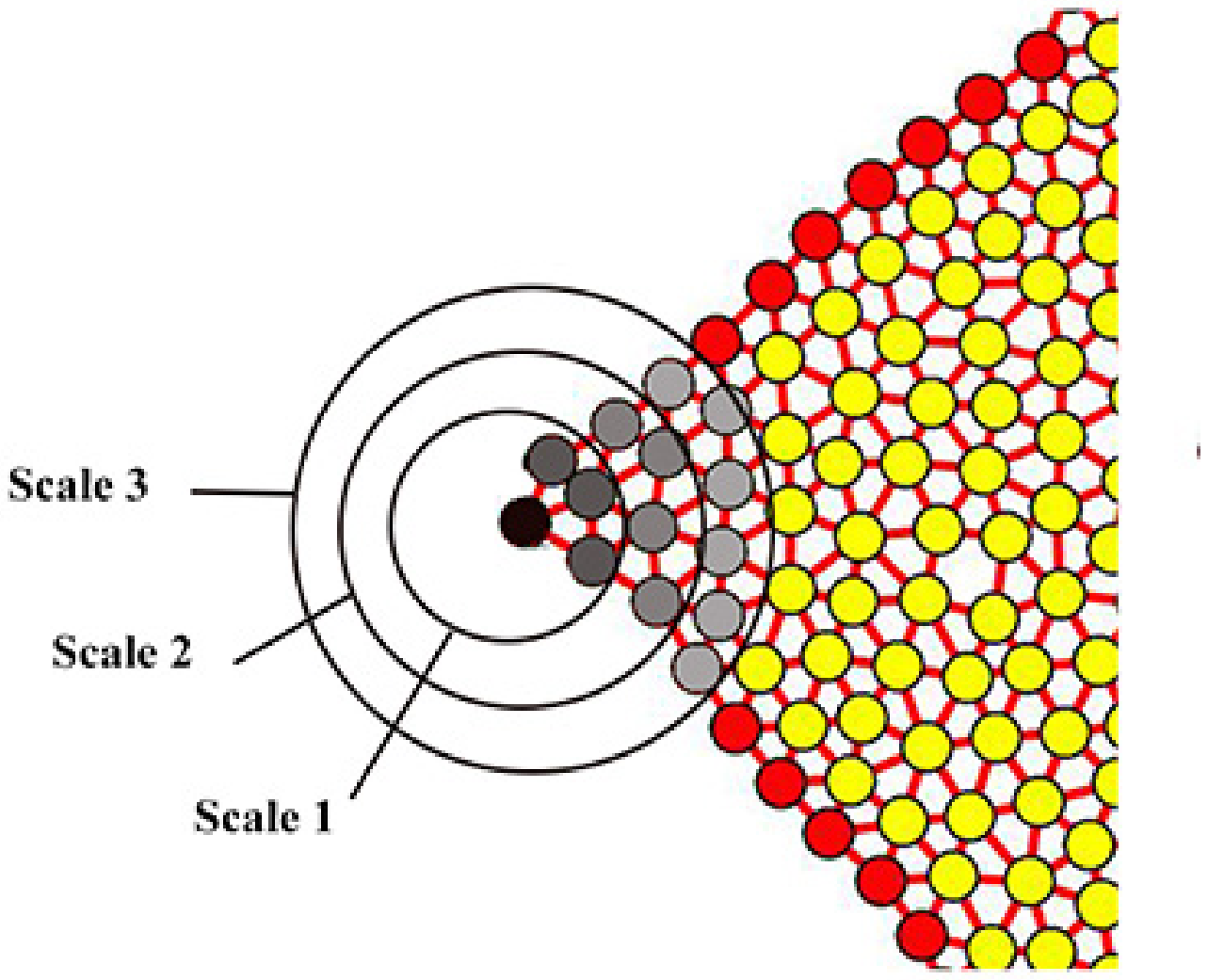}
\centering
\caption{\label{fig:scale_gng}The scales 1-3 around the black vertex are shown and the vertices with the same distances 1-3 from black vertex are specified in the same color.}
\end{figure}

 \item \textbf{Perimeter $(P)$}:

 For each outer boundary vertex $u_i$, perimeter counts the number of vertices that are in $\bar{D}_j(u_i)$ for $ j\in \lbrace 1,...,m_1 \rbrace $. Small values of $j$ describe local properties while larger values represent global properties of the shape.\\
Figure \ref{fig:feature1and2}-a shows an example of this feature.
\item \textbf{Boundary-in-disk $(B)$}:

 The number of outer boundary vertices inside $D_j(u_i)$ for each outer boundary vertex $u_i$ where $j \in \lbrace 1,...,m_2 \rbrace $ is called boundary-in-disk. If the boundary has peaks and troughs inside the disk, this number is bigger, so this feature can keep the shape of the boundary, if it is computed for different radii (see figure \ref{fig:feature1and2}-b).

 \item \textbf{Convex hull-area $(CH)$}:\\
For each outer boundary vertex $u_i$ and integer $j\in \lbrace 1,2,...,m_3\rbrace$, let $S$ be the convex hull of vertices in $D_j(u_i)$ that are on the outer boundary of $G$. The number of vertices of $D_j(u_i)$ that are inside $S$ is called convex hull-area.

 \item \textbf{Distance-to-center $(C)$}:
For vertex $u_i$, distance-to-center $(C)$ is measured as the ratio of distances between $u_i$ and the center of $G$ and $j$, $j \in \lbrace 1,...,m_4 \rbrace$. \\

 \vspace{-0.5cm}
\begin{figure*}[!htb]
\includegraphics[scale=.6]{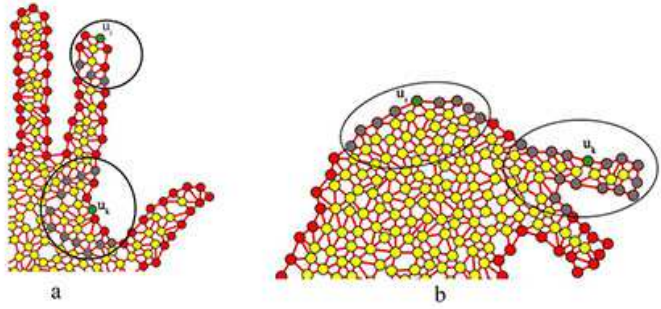}
\centering
\caption{\label{fig:feature1and2}a) An example of feature $P$ for two different vertices $u_i$ and $u_k$ when $j=3$. Here $P_i=3$ and $P_j=11$. The vertices in $\bar{D}_3(u_i)$ or $\bar{D}_3(u_k)$ are colored in gray. b) An example of feature $B$ for two vertices $u_i$ and $u_k$ when $j=4$. Here $B_i=8$ and $B_k=15$. These vertices are colored in gray.}
\end{figure*}

 \begin{figure}[!htb]
\includegraphics[scale=.4]{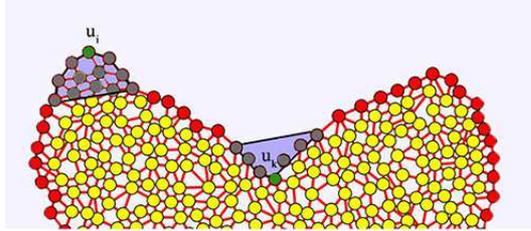}
\centering
\caption{\label{fig:feature_CH} An example of $CH$ at two vertices $u_i$ and $u_k$ for $j=3$. Here $CH_i=11$ and $CH_k=6$. These vertices are colored in gray.}
\end{figure}
\end{itemize}

 Yang et al. in \cite{YangInvariant2016} considered circles with different radii on each boundary point. They defined a major zone as the connected region of the object which is inside the circle and contains the center of it.
They computed features including area and length of the boundary segment of major zone, and the distance between the center of circle and the center of major zone. To determine the scale number $m$ (maximum radius), they follow this condition:
if the average difference of features between neighboring scales $m$ and $m+1$ is less than a threshold, The scale $m$ is enough \cite{YangInvariant2016}. We applied a similar approach to determine $m_1,m_2,m_3$ and $m_4$.

\section{Similarity measuring and matching }
\label{sec:Similarity measureing}

 We apply Dynamic programming (DP) to find an optimal matching between two objects and consider the matching cost as the dissimilarity value of them. DP is an effective method for contour matching with high accuracy.
Let $\lbrace p_1,p_2,...,p_n \rbrace$ and $\lbrace q_1,q_2,...,q_m\rbrace$ be the boundary sequence of two objects $A$ and $B$ respectively. A matching $\pi$ from $A$ to $B$ is defined as a mapping from $\lbrace p_1,p_2,...,p_n \rbrace$ to $\lbrace q_1,q_2,...,q_m\rbrace$ where $p_i$ is mapped to $q_{\pi(i)}$ if $\pi(i)\neq 0 $ and otherwise $p_i$ remains unmapped. The cost of $\pi$ is defined as $\sum_{i=1}^{n}c(p_i,q_{\pi(i)})$, where $c(p_i,q_{\pi(i)})$ equals the Euclidean distance between feature vectors $p_i$ and $q_{\pi(i)}$. DP computes a matching with minimum cost between sequences of boundary points of two objects.

\begin{figure}[!htb]
\includegraphics[scale=.3]{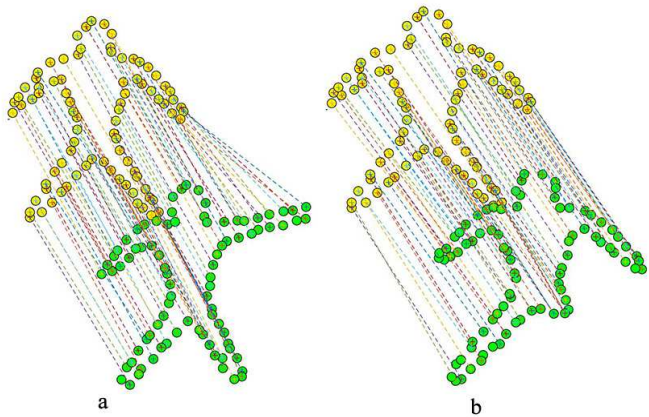}
\centering
\caption{\label{fig:matching} A matching between outer boundary vertices of two objects. Dashed lines show the mapping. a) shows a matching with articulation and b) shows a matching with occlusion.}
\end{figure}

 We compute outer boundary vertices of the GNG graph and consider them as boundary points in DP. Figure \ref{fig:matching} shows the examples of matching between two different objects.

 \section{Comparison}\label{sec:Comparison}
To evaluate our method, we provide experimental comparison with the state-of-the-art methods on challenging benchmark datasets Kimia\textquotesingle s 99, Kimia\textquotesingle s 216 \cite{Sebastian2004}, Tetrapod \cite{Latecki2000}, Tari56 \cite{Tari56} and Articulated dataset \cite{Ling}. These datasets are well-known in the object recognition area and have been used by various methods to evaluate \cite{Belongie2007,Ling,Wang2012,YangInvariant2016,Yang2016}.
The datasets include the challenges scale variation, rotation, articulation, intra-class variation, and missing parts. The retrieval rate is measured by the so-called \textit{bull \textquotesingle s eyes} score.
It measured the similarity between an object and all other objects.
The number of objects belonging to the same class among the top 1 to n most similar objects is counted (parameter n is different in various datasets). Every object in the dataset is used as a query, and the retrieval result for the total dataset is computed by averaging among all objects \cite{Wang2012}.

\begin{figure}[!htb]
\includegraphics[scale=1.2]{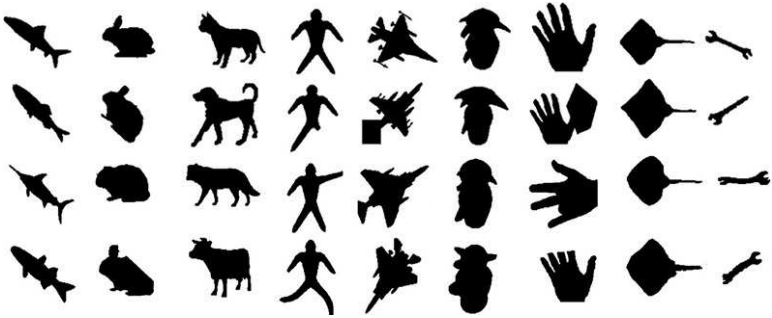}
\centering
\captionsetup{justification=centering}
\caption{\label{fig:Kimia99} Some images of Kimia \textquotesingle s 99 dataset.}
\end{figure}

\begin{table*}[!htb]
\begin{center}
\caption{Exprimental comparison of our method with state-of-the-art methods on Kimia\textquotesingle s 99}\label{table:Kimia99}
\scalebox{1}{
\begin{tabular}{ l c c c c c c c c c c }
\hline \textbf{Kimia\textquotesingle s 99 }&\textbf{ 1st} & \textbf{2nd} & \textbf{3rd }&\textbf{ 4th} & \textbf{5th}& \textbf{6th }&\textbf{7th} &\textbf{ 8th}&\textbf{ 9th} & \textbf{10th}\\
\hline
SC \cite{Belongie2007} & 99 & 97 & 91 & 88 & 84 &83 & 76 &76 & 68 & 62 \\
IDSC \cite{Ling} & 99 & 97 & 92 & 89 & 85 & 85 & 76 & 75 & 63 & 53 \\
Path similarity\cite{Latecki2000}& 99 & 99& 99&99 &96 &97 &95&93 &89 &73\\
Invariant multi-scale \cite{YangInvariant2016} & 99 &99 &99& 99& 98 &97& 95& 94 &90 & 83\\
Hierarchical skeletons \cite{Yang2016}& 99 & 99 & 99 & 96 & 94 &95 & 91 & 89 & 85 & 77 \\
\textbf{Ours }& 99 & 98 & 96 & 96 & 96 & 94 & 95 & 90 & 88 & 86 \\
\hline
\end{tabular}
}
\end{center}
\end{table*}

 \begin{figure}[!htb]
\includegraphics[scale=.6]{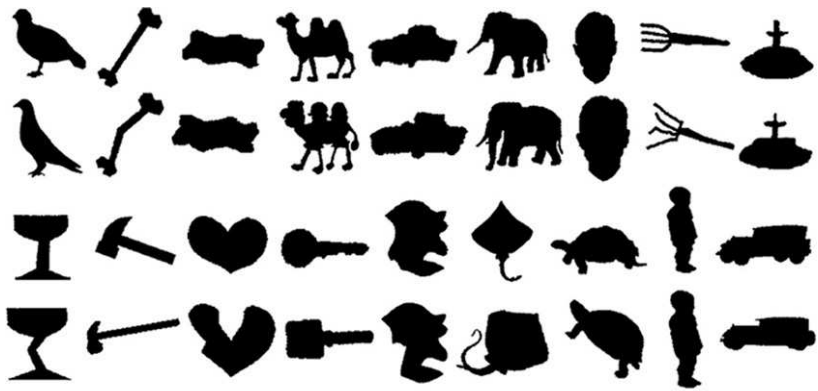}
\centering
\caption{\label{fig:Kimia216} Some images of Kimia \textquotesingle s 216 dataset \cite{Yang2016}.}
\end{figure}

\subsection{Kimia\textquotesingle s dataset}
The Kimia\textquotesingle s dataset is a widely used benchmark dataset in shape classification \cite{Sebastian2004}. It contains Kimia\textquotesingle s 25, Kimia\textquotesingle s 99 and Kimia\textquotesingle s 216 datasets. The Kimia\textquotesingle s 25 is a small dataset and contains only 25 objects, so we consider Kimia\textquotesingle s 99 and Kimia\textquotesingle s 216 to evaluate our method.\\
\textbf{Kimia\textquotesingle s 99}: Kimia\textquotesingle s 99 has 99 images grouped into 9 different classes that occlusions, articulations and missing parts occur in this data set (see figure \ref{fig:Kimia99}).
The retrieval result is summarized as the number of top 1 to top 10 most similar objects of correct category (the best possible retrieval rate is 99). The comparison of results on Kimia\textquotesingle s 99 is reported in table \ref{table:Kimia99}. Our performance is comparable on Kimia\textquotesingle s 99.\\
\textbf{Kimia\textquotesingle s 216}: Kimia\textquotesingle s 216 includes 216 images from 18 categories. Figure \ref{fig:Kimia216} shows some objects of Kimia\textquotesingle s 216.
Our results on Kimia\textquotesingle s 216 are reported in table \ref{table:Kimia216}. It shows our method has considerable performance among other approaches.
\begin{table*}[!htb]
\begin{center}
\caption{Experimental comparison of our method with state-of-the-art methods on Kimia\textquotesingle s 216}\label{table:Kimia216}
\scalebox{1}{
\begin{tabular}{ l c c c c c c c c c c c }
\hline \textbf{Kimia\textquotesingle s 216 }&\textbf{ 1st} & \textbf{2nd} & \textbf{3rd }&\textbf{ 4th} & \textbf{5th}& \textbf{6th }&\textbf{7th} &\textbf{ 8th}&\textbf{ 9th} & \textbf{10th}& \textbf{11th}\\
\hline
SC \cite{Belongie2007} & 204 & 199 & 192 & 187 & 185 &181 & 175 &166 & 160 & 163 & 155\\
IDSC \cite{Ling}& 216 & 198 &189 & 176 & 167 &156 & 136 & 130 & 122 & 118 &108\\
Path similarity \cite{Latecki2000}& 216 &216 & 215& 216& 213& 210& 210& 207 &205 & 191& 177\\
Invariant multi-scale \cite{YangInvariant2016} & 216 & 216 & 214 & 210 & 207 & 207 & 201 & 194 & 188 & 182 & 163 \\
Hierarchical skeletons \cite{Yang2016}& 216 & 216 & 213 & 212 &209 &197 &196 & 192 & 193 &172 & 169\\
\textbf{Ours }& 216 & 216 & 214 & 212 & 211 & 206 & 210 & 200 & 195 & 186 & 163\\

 \hline
\end{tabular}
}
\end{center}
\end{table*}

\begin{figure*}[!htb]
\includegraphics[scale=.45]{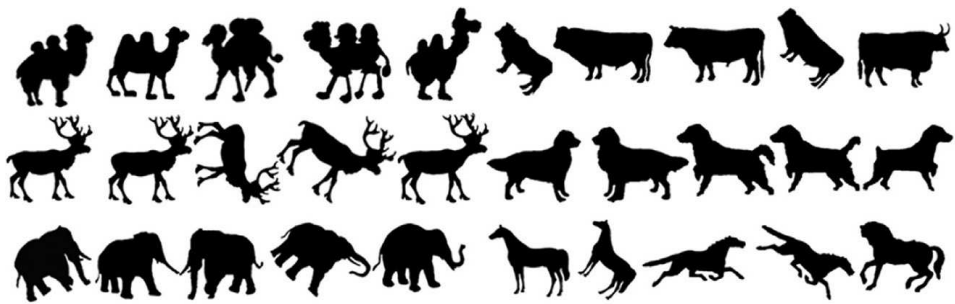}
\caption{\label{fig:tetrapod} Some images of Tetrapod dataset \cite{Yang2016}.}
\end{figure*}

\begin{table*}[!htb]
\centering
\caption{Exprimental comparison of our method with state-of-the-art methods on Tetrapod dataset}\label{table:Tetrapod}
\scalebox{1}{
\begin{tabular}{ l c c c c c c c c c c }
\hline \textbf{Tetrapod }&\textbf{ 1st} & \textbf{2nd} & \textbf{3rd }&\textbf{ 4th} & \textbf{5th}& \textbf{6th }&\textbf{7th} &\textbf{ 8th}&\textbf{ 9th} & \textbf{10th}\\
\hline
SC \cite{Belongie2007} & 100 & 80 &70 & 53 & 53 & 51 & 40 & 28 & 27 & 27\\
IDSC \cite{Ling} & 120 & 118 &106 & 101 & 90 &83 & 77 & 69 & 70 & 56\\
Path similarity\cite{Latecki2000} & 120 & 109 & 101 & 98 & 81 & 78 & 68 & 66 & 65 & 59\\
Hierarchical skeletons \cite{Yang2016}& 120 & 118 & 106 &100 & 95 & 90 & 84 & 71 & 83 & 81 \\
\textbf{Ours }&\textbf{120}&\textbf{119} &\textbf{114} &\textbf{113} &\textbf{ 112}&\textbf{ 110}&\textbf{ 111}&\textbf{99} &\textbf{ 99} &\textbf{95} \\
\\
\hline
&\textbf{11th} & \textbf{12th}& \textbf{13th}& \textbf{14th}& \textbf{15th}&\textbf{16th}& \textbf{17th}& \textbf{18th}& \textbf{19th}& \textbf{20th}\\
\hline
SC \cite{Belongie2007} & 29 & 27 & 25 & 32 & 32 & 23 & 31 &26 & 20 & 28\\
IDSC \cite{Ling}& 57& 45& 38 &29 &41 & 35 & 26 &27 &30& 21\\
Path similarity \cite{Latecki2000}&59 & 49 & 50 & 42 & 43 & 35 & 39 & 31 &33 & 36 \\
Hierarchical skeletons \cite{Yang2016} & 68 & 73 & 67 & 77 &68 & 67 & 60 & 51 & 56 &43 \\
\textbf{Ours} &\textbf{ 83} &\textbf{83}& \textbf{78}&76 & 67& 61 &\textbf{62}&\textbf{52}& 44 &\textbf{ 45}\\
\hline
\end{tabular}
}
\end{table*}

 \subsection{Tetrapod dataset}
Tetrapod contains 120 images from similar tetrapod animals and grouped into 6 classes deer, camel, elephant, cattle, dog, and horse \cite{Latecki2000} (see figure \ref{fig:tetrapod}). The high intra-class similarity of the dataset is very challenging. As shown in table \ref{table:Tetrapod}, Our method achieved considerably better performance than other approaches with the bull \textquotesingle s eyes score.

\begin{figure}[!htb]
\includegraphics[scale=.9]{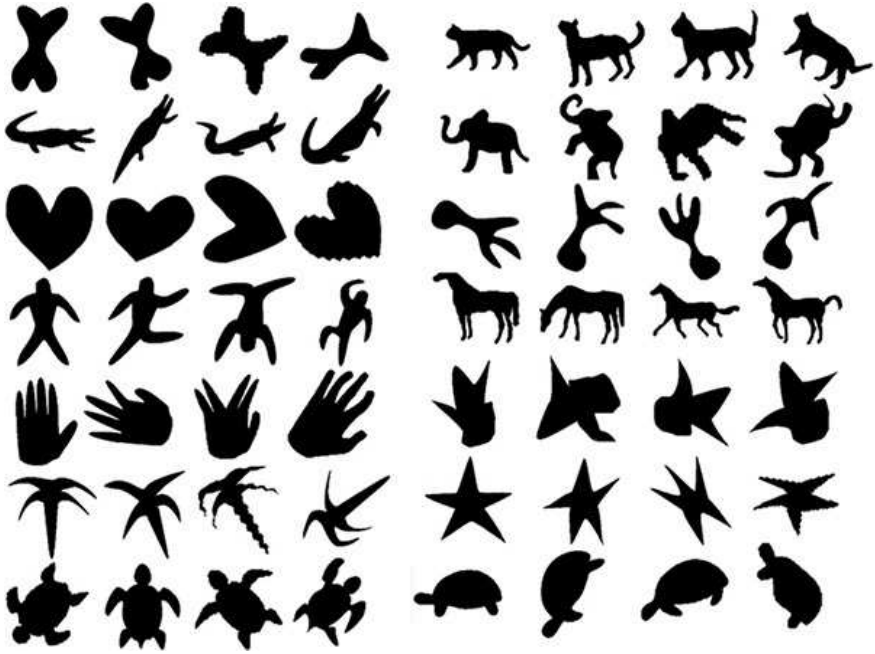}
\centering
\caption{\label{fig:tari56} All images of tari56 dataset \cite{Yang2016}.}
\end{figure}

 \subsection{Tari56 dataset}
Tari56 is introduced for evaluating on non-rigid objects by Tari \cite{Tari56}. It includes 56 similar articulated shapes grouped in 14 classes with 4 shapes in any class (see figure \ref{fig:tari56}). Yang et al. implemented different source code methods on Tari56 and presented an experimental comparison with them \cite{Yang2016}. We use the reported compared results in
 \cite{Yang2016} to evaluate our method. The results of comparison on Tari56 with the bull \textquotesingle s eyes score are listed in table \ref{table:Tari56}. It shows that our results are significantly better among all other approaches.
\begin{table}[!htb]
\begin{center}
\caption{Experimental comparison of our method with state-of-the-art methods on Tari56 dataset}\label{table:Tari56}
\scalebox{.9}{
\begin{tabular}{ l c c c c }
\hline
\textbf{Tari56 dataset }&\textbf{ 1st} & \textbf{2nd} & \textbf{3rd }&\textbf{ 4th} \\
\hline
SC \cite{Belongie2007} & 52 & 17 & 10 & 10 \\
IDSC \cite{Ling}& 56 & 46 & 37 & 28 \\
Path Similarity \cite{Latecki2000} & 56 & 49 & 44 & 40\\
Hierarchical skeletons \cite{Yang2016}& 56 & 51 & 50 & 33\\
\textbf{Ours}& \textbf{56 }& \textbf{ 55}& \textbf{ 53} &\textbf{ 53}\\
\hline
\end{tabular}
}
\end{center}
\end{table}

\begin{figure*}[!htb]
\includegraphics[scale=.7]{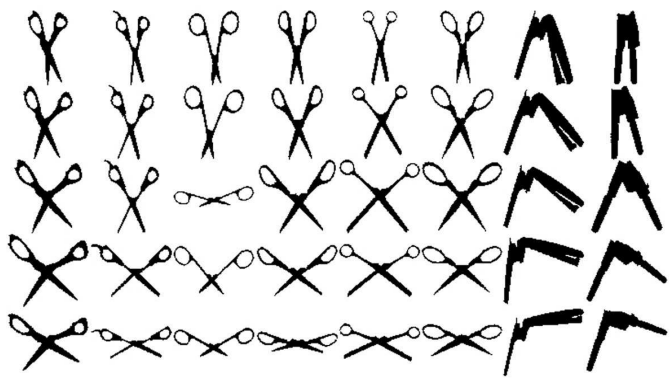}
\centering
\caption{\label{fig:articulated} All images of articulated dataset \cite{Ling}.}
\end{figure*}

 \subsection{Articulated dataset}
Articulated dataset provided by Linge and Jacobs \cite{Ling} contains 40 images from 8 different classes and each class has 5 articulated shapes to different degrees.
In addition to articulation, high similarity between different classes of scissors is challenging. Figure \ref{fig:articulated} shows all images of this dataset. The results of comparison are listed in table \ref{table:articulated}. Our results on this dataset are comparable to other methods.
\begin{table}[!htb]
\begin{center}
\caption{Experimental comparison of our method with state-of-the-art methods on articulated dataset}\label{table:articulated}
\scalebox{.85}{
\begin{tabular}{ l c c c c }
\hline
\textbf{Articulated dataset }&\textbf{ 1st} & \textbf{2nd} & \textbf{3rd }&\textbf{ 4th} \\
\hline
SC \cite{Belongie2007} & 20 & 10 & 11 & 5 \\
IDSC \cite{Ling}& 40 & 34 & 35 & 27 \\
Height functions \cite{Wang2012} & 38 & 35 & 28 & 19\\
HCNC \cite{Jia2016} & 40 & 38 & 29 & 22\\
Invariant multi-scale \cite{YangInvariant2016} & 40 & 37& 35& 30 \\
\textbf{Ours}& 39 & 37 & 34 & 26\\
\hline
\end{tabular}
}
\end{center}
\end{table}

\subsection{Analysis properties of our method}

\begin{figure*}[htb]
\begin{center}
\makebox[\linewidth]{
\includegraphics[width=.85 \linewidth]{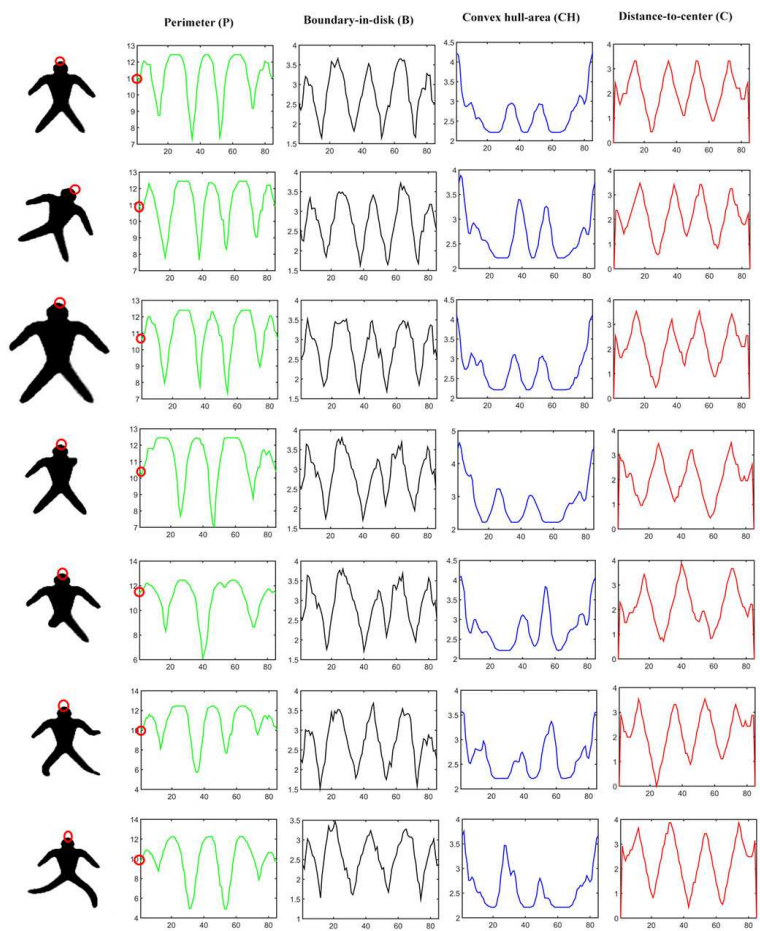}
}
\caption{\label{fig:invariant}The diagram of features perimeter $(P)$, boundary-in-disk $(B)$, convex hull-area $(CH)$ and distance-to-center $(C)$.}
\end{center}
\end{figure*}

 To study the behavior of different features in the presence of noise, articulation, scale, rotation, occlusion an intra-class variation, we made an experimental study as follows.
We selected 5 images from a class of Kimia\textquotesingle s 216, then we create new images with changing the scale and rotation of these images, afterwards we plot the features of objects to compare their diagrams. Figure \ref{fig:invariant} shows these diagrams. In order to make the comparison easier, we start the boundary list from the red circle shown in each figure. Columns 1 to 4 present the plot of features Perimeter, Boundary-in-disk, Convex hull-area and Distance-to-center, respectively. The values shown in these diagrams are the average of values in different scales.

 \subsubsection{ Scale}
We represent the image of objects by graphs with a fixed number of vertices. Therefore objects with different scales are represented as graphs with the same size, so our method is invariant to scale. Rows 1 and 3 in figure\ref{fig:invariant} show the plot of different features for two different scales of an image. As these plots show, the features dot not change seriously when the scale changes.
\subsubsection{Rotation}
Graphs have interesting characteristics which can play an important role in object recognition: graphs do not change by rotation, so, our method is robust to rotation. Experimental studied approve this statement. Rows 1 and 2 in figure \ref{fig:invariant} show the same object with different rotations. Comparing columns 1 to 4 in these rows show that the corresponding features are very similar.
\subsubsection{Occlusion}
A significant property of GNG graph is its low dimensionality. It models objects with low dimensional subspaces which reflect the topological structure of them. Since GNG graph preserves the topological properties of the image, occlusion and missing parts have little effect on this graph.
Rows 4 and 5 in figure \ref{fig:invariant} show images that include occlusion. We see that the missing parts do not destruct the features of other parts and matching other parts help correct object recognition. Figure \ref{fig:matching}-b shows the matching when the missing parts have occurred.
 \subsubsection{ Intra-class variation}
Rows 6 and 7 of figure \ref{fig:invariant} represent the examples of intra-class variations. We see that the corresponding columns 1, 2 and 4 are very similar. The general structure of the features such as the main peaks in the diagrams are similar.

 \subsubsection{Articulation}
\begin{figure}[!htb]
\captionsetup{justification=centering}
\includegraphics[scale=.17]{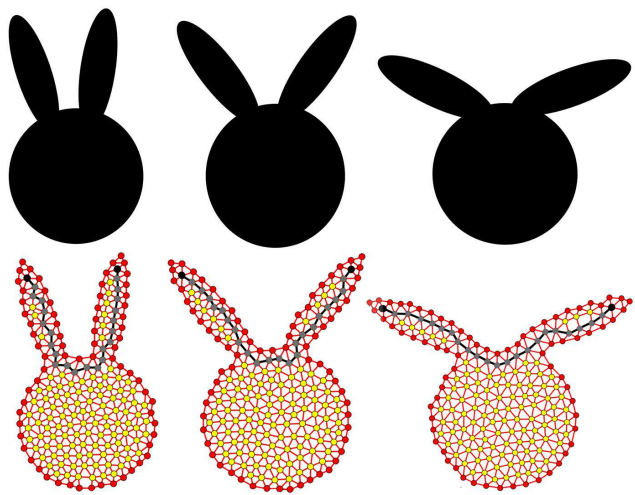}
\centering
\caption{\label{fig:articulation_gng}
First row shows objects with different articulations and second row shows their GNG graph and inner paths between black vertices.}
\end{figure}

Articulation is an other challenge in object recognition. We evaluated our method on Articulated dataset.
In this part, we take a look to different features in articulated objects. Some features like the distance in graph do not depend on the coordinate and hence do not change in articulation. Figure \ref{fig:articulation_gng} shows objects with different articulations and their corresponding GNG graph and the shortest path between two vertices of it. The length of inner distances is similar without using normalization.

\begin{figure*}[!htb]
\vspace*{2cm}
\centering
\includegraphics[scale=.4]{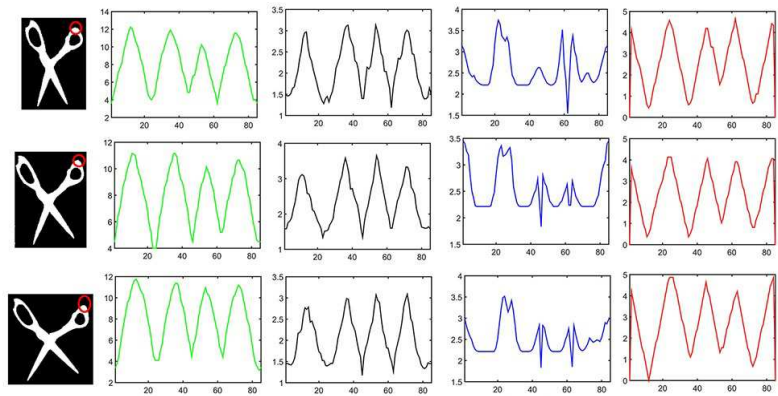}
\caption{\label{fig:articulation_invariant} The objects from the same class with different articulations and the plot of correspondence features. Red circles show the beginning position of boundary.}

\end{figure*}
 In a second observation, we take a look at the plot of different features in some images with articulation.
Figure \ref{fig:articulation_invariant} shows these plots.


 \subsubsection{Robustness against noise}
\begin{figure*}[!htb]

\centering
\includegraphics[scale=1]{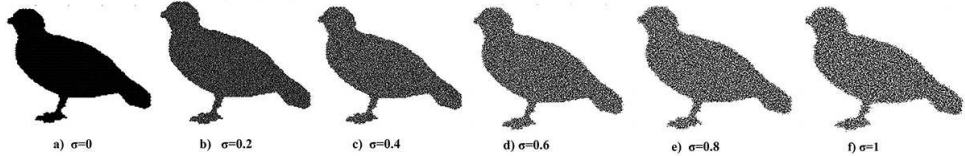}
\caption{\label{fig:noisey_object} The same object with Gaussian noises in different standard deviations. }
\end{figure*}

Noise is a common problem in real world applications.
In this section we select an image from Kimia\textquotesingle s 216 dataset and add Gaussian noise with zero mean and standard deviation $\sigma$ as 0.2, 0.4, 0.6, 0.8 and 1 to all pixels on each image in both x and y directions.
These images are shown in figure \ref{fig:noisey_object}.

\begin{figure*}[htb]
\centering
\includegraphics[scale=.17]{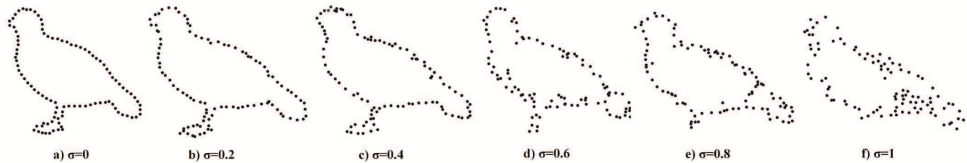}
\caption{\label{fig:noise_height} The sample points of outer boundary extracted by \cite{Wang2012}. }
\end{figure*}

 Finding the boundary in the images with noise is a challenging problem. We used the algorithm in \cite{Wang2012} for these images and show the result in figure \ref{fig:noise_height}.

\begin{figure}[htb]
 \centering
\includegraphics[scale=.2]{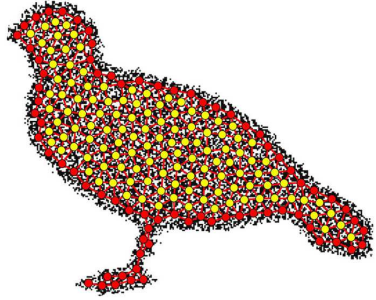}
\caption{\label{fig:noise_sigma_1} The GNG graph of object with Gaussian noise $\sigma=1$. }
\end{figure}

 As figure \ref{fig:noise_height} shows, the boundary can be very messy therefor the features extracted from boundary do not work properly in object recognition. But in figure \ref{fig:noise_sigma_1} shows despite the loss of image pixels, GNG graph models object as well and the outer boundary of object is detected with high accuracy.

\begin{figure*}[!htb]
 \vspace{4cm}
 \centering
\includegraphics[scale=.2]{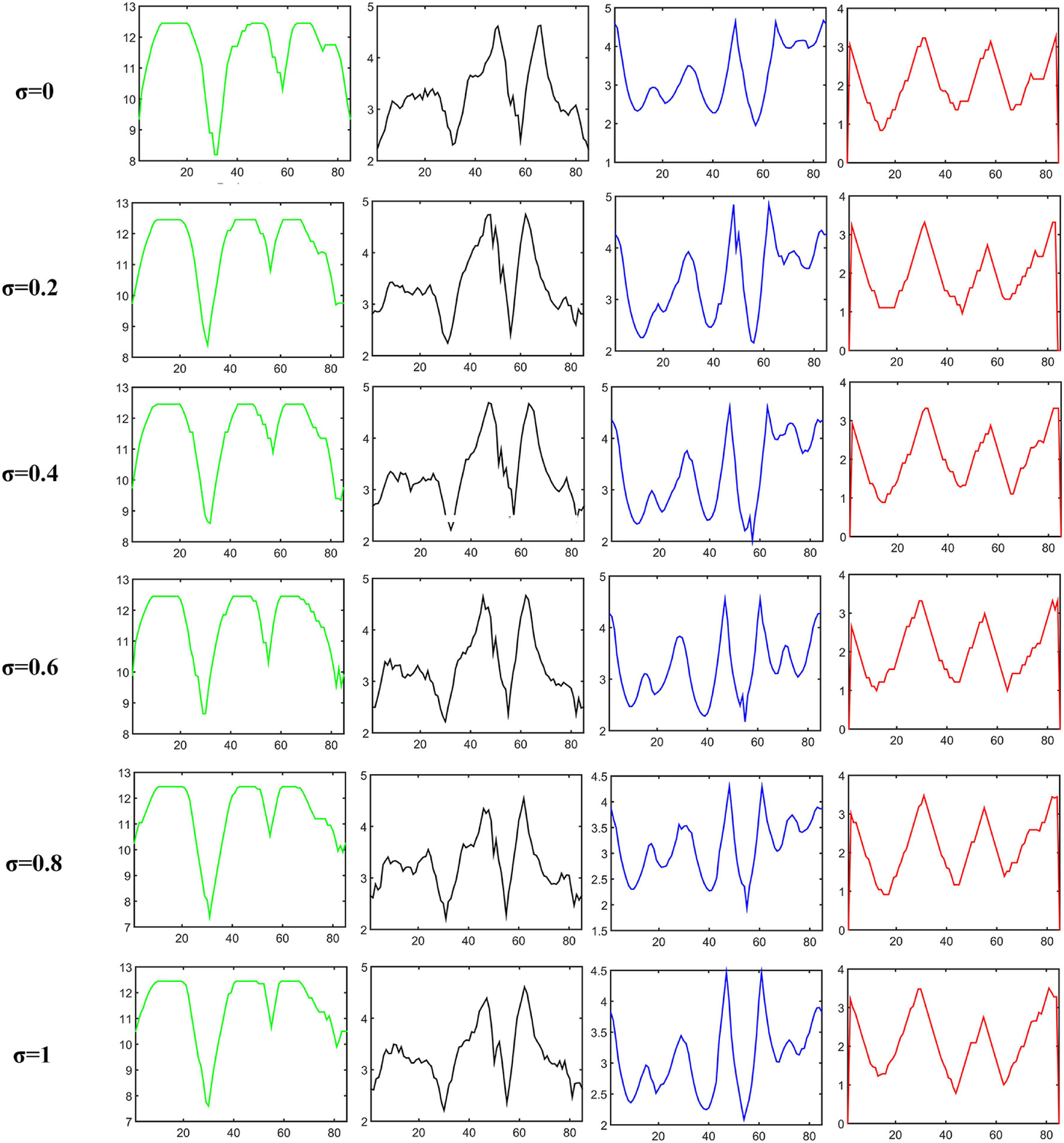}
\caption{\label{fig:invariant_noise} The features of the same object with Gaussian noises in different standard deviations (figure \ref{fig:noisey_object}).
}
\end{figure*}

 Finally, we plot the features of noisy images of figures \ref{fig:noisey_object} (see figure \ref{fig:invariant_noise}). We also applied Gaussian noises on all images in Kimia's 216 and reported the results of recognition algorithm in table \ref{table:Kimia216_noise}.

 \begin{table*}[!htb]
\begin{center}
\caption{Experimental results on Kimia\textquotesingle s 216 with Gaussian noises in different standard deviations }\label{table:Kimia216_noise}
\scalebox{1}{
\begin{tabular}{ l c c c c c c c c c c c }
\hline \textbf{Kimia\textquotesingle s 216 }&\textbf{ 1st} & \textbf{2nd} & \textbf{3rd }&\textbf{ 4th} & \textbf{5th}& \textbf{6th }&\textbf{7th} &\textbf{ 8th}&\textbf{ 9th} & \textbf{10th}& \textbf{11th}\\
\hline
$\sigma=0$ & 216 & 216 & 214 & 212 &211 & 206 & 210 & 200 & 195 &186 & 163 \\
$\sigma=0.2$& 215 & 215 & 216 & 210 & 208 & 210 & 209 & 202 & 194 & 185 & 161\\
$\sigma=0.4$ & 215 & 216 & 213 & 211 & 211 & 211 & 209 & 198 & 196 & 182 & 158\\
$\sigma=0.6$ & 216 & 215 & 216 & 212 & 209 & 210 & 208 & 198 & 200 & 189 & 164\\
$\sigma=0.8$ & 215 & 216 & 214 & 211 & 211 & 210 & 205 & 196 & 195 & 188 & 151\\
$\sigma=1$ &214 & 213& 212& 210 &207 & 204 &205 &197 & 191 & 186 &159\\
\hline
\end{tabular}
}
\end{center}
\end{table*}

 \subsection{Our contribution}
Pixel-based methods are dependent on the geometry and nature of the pixels, so the destruction of pixels such as small inner holes, the noise of the boundary, and the dispersion of pixels mainly bother them and reduce their ability.
Most existing pixel-based methods are contour-based and are dependent on other methods to extract the contour of objects, while extracting the contour of objects in a noisy space is a challenging problem.
To avoid the limitation of pixel-based methods, this paper introduces a more flexible method based on the graph. We show the ability of graphs to shape representation.
Our performance is similar and comparable to well pixel-based methods. In fact, we introduce a new space for object recognition with significant properties which can be improved by other researches.
Our main contributions are:
\begin{itemize}
\item
Object representation in graph space.
\item
Outer boundary extraction without dependence on pixel-based algorithms.
\item
Definition of new graph-based features that capture the topological and geometrical properties of the object.
\item
High flexibility against articulation.
\item
Considerable stability against noises and small inner holes.
\item
No need to normalize the features, which reduces computational errors.
\item
Tracking ability of the behavior of nodes in online images \cite{Sun2017}.
\end{itemize}
Also, our method preserves the advantage of other methods such as stability against translation, sale, rotation, occlusion and missing parts.
We show many pixel-based features can be defined with graphs. We defined features similar with Invariant multi-scale \cite{YangInvariant2016} in graph domains without the limitation of pixel against noises.
\section{Conclusion and future work}\label{sec:conclusion}
This work provides a novel graph-based approach for object recognition.
In this paper, we study the role of graph in image representation.
We model objects with GNG graphs and use the coordinate and relation of vertices to extract topological features that are robust to noise, rotation, scale variation, and articulation.
These features describe global and local properties of an object. The proposed method uses DP to measure the similarity between objects. We evaluate our methods on different known benchmarks. We analyze our method to articulation, scale, occlusion, and rotation. The experimental results indicate comparability of our method with state-of-the-art methods.
Also, we evaluate our method in the presence of noises. The results show the high ability of our method against noises.
In this work, we showed that object recognition can be improved using graphs for image representing and extracting topological features.
Hand gesture and American sign language recognition are interesting problems because of high intra-class similarities between different signs. Studying the performance of our method on these problems can be interesting.


\end{document}